\documentclass[a4paper]{article}

\usepackage{INTERSPEECH2021}
\usepackage{todonotes}
\usepackage{subcaption}

\title{Visualizing Automatic Speech Recognition \\-- Means for a Better Understanding?}
\name{Karla Markert$^{*,1,2}$, Romain Parracone$^{*,2}$, Mykhailo Kulakov$^{*,2}$, \\Philip Sperl$^{1,2}$, Ching-Yu Kao$^{1,2}$, Konstantin Böttinger$^{1}$}
\address{
  $^1$Fraunhofer AISEC, Germany; 
  $^2$Technical University Munich, Germany\\
  $^*$authors contributed equally}
\email{\{karla.markert, philip.sperl, ching-yu.kao, konstantin.boettinger\}@aisec.fraunhofer.de \\ \{romain.parracone, michael.kulakov\}@tum.de}

\begin{document}

\maketitle
\begin{abstract}
  Automatic speech recognition (ASR) is improving ever more at mimicking human speech processing. 
  The functioning of ASR, however, remains to a large extent obfuscated by the complex structure of the deep neural networks (DNNs) they are based on. 
  In this paper, we show how so-called attribution methods, that we import from image recognition and suitably adapt to handle audio data, can help to clarify the working of ASR. 
  Taking DeepSpeech, an end-to-end model for ASR, as a case study, we show how these techniques help to visualize which features of the input are the most influential in determining the output. 
  We focus on three visualization techniques:
  Layer-wise Relevance Propagation (LRP), Saliency Maps, and Shapley Additive Explanations (SHAP). 
  We compare these methods and discuss potential further applications, such as in the detection of adversarial examples.
\end{abstract}



  



\section{Introduction}
\begin{figure*}[h!]
    \centering
    \includegraphics[width=0.9\textwidth]{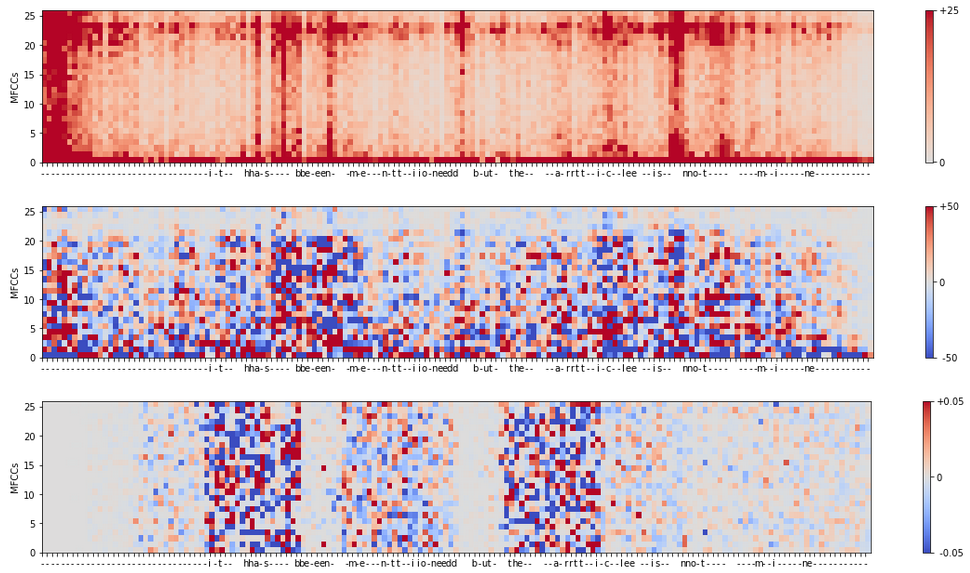}
    \caption{Different visualizations of the benign example ``it has been mentioned but the article is not mine''.
    (Top) The \textbf{Saliency Map} has the highest attributions at low and high frequencies.
    (Middle) In the \textbf{LRP}, the lowest ten MFCCs are the most attributing ones.
    (Bottom) For the \textbf{SHAP}, the attributions show instead a more clustered-in-time behaviour while being equally distributed along the different MFCC values.}
    \label{fig:AttributionMethodsComparison}
\end{figure*}

DNNs find nowadays applications across a rich variety of domains, including computer vision, language translation, and speech or recognition.
Alongside the increase of available computing power, DNNs' architectures have became extremely involved and may consist of up to millions of trainable parameters. 
Such complexity makes the important task of understanding why a model outputs a certain prediction challenging \cite{zhang2019towards}. 

The pursuit of understanding DNNs' has given rise to the new field of interpretable (or explainable) machine learning \cite{gilpin2018explaining}. 
In this context, a class of interpretability algorithms called attribution-based methods has been developed \cite{ancona2017towards,sundararajan2017axiomatic} to evaluate the attributions of certain features of the DNN's input, that is, their effect on the output. 
In the past, these techniques have been mainly applied to image object classification and natural language processing \cite{ancona2021towards, zhou2021feature} and the question whether similar techniques can provide insights in ASR remains open. 
In this paper, we address this question focusing on the example of DeepSpeech and discuss how attribution-based approaches can help to enable model transparency.

Indeed, a better understanding of the DNNs is crucial to allow the design of ASR with higher and higher accuracy. 
Furthermore, a clearer visualization of DNNs would potentially allow developers to design more secure ASR.
With the involvement of machine learning in more and more aspects of our daily life, the discussion around security has been brought to the forefront.
In certain applications, like voice control systems for cars, the security of the ASR system becomes critical even for the user's physical safety.

As a specific example, a better understanding of DNNs could help to prevent so-called adversarial examples.
These intentionally perturbed inputs trick the attacked models into misclassifications. 
Indeed, explanation methods have already proven helpful in understanding adversarial examples and dataset-related issues in the context of image analysis \cite{gu2019saliency, etmann2019connection}, and have the potential to provide similar insights in support of recent analyses on audio adversarial examples for ASR \cite{carlini2018audio,qin2019imperceptible,dorr2020towards,hu2019adversarial,abdullah2020sok}.

\begin{figure}[h!]
    \centering
    \includegraphics[width=0.45\textwidth]{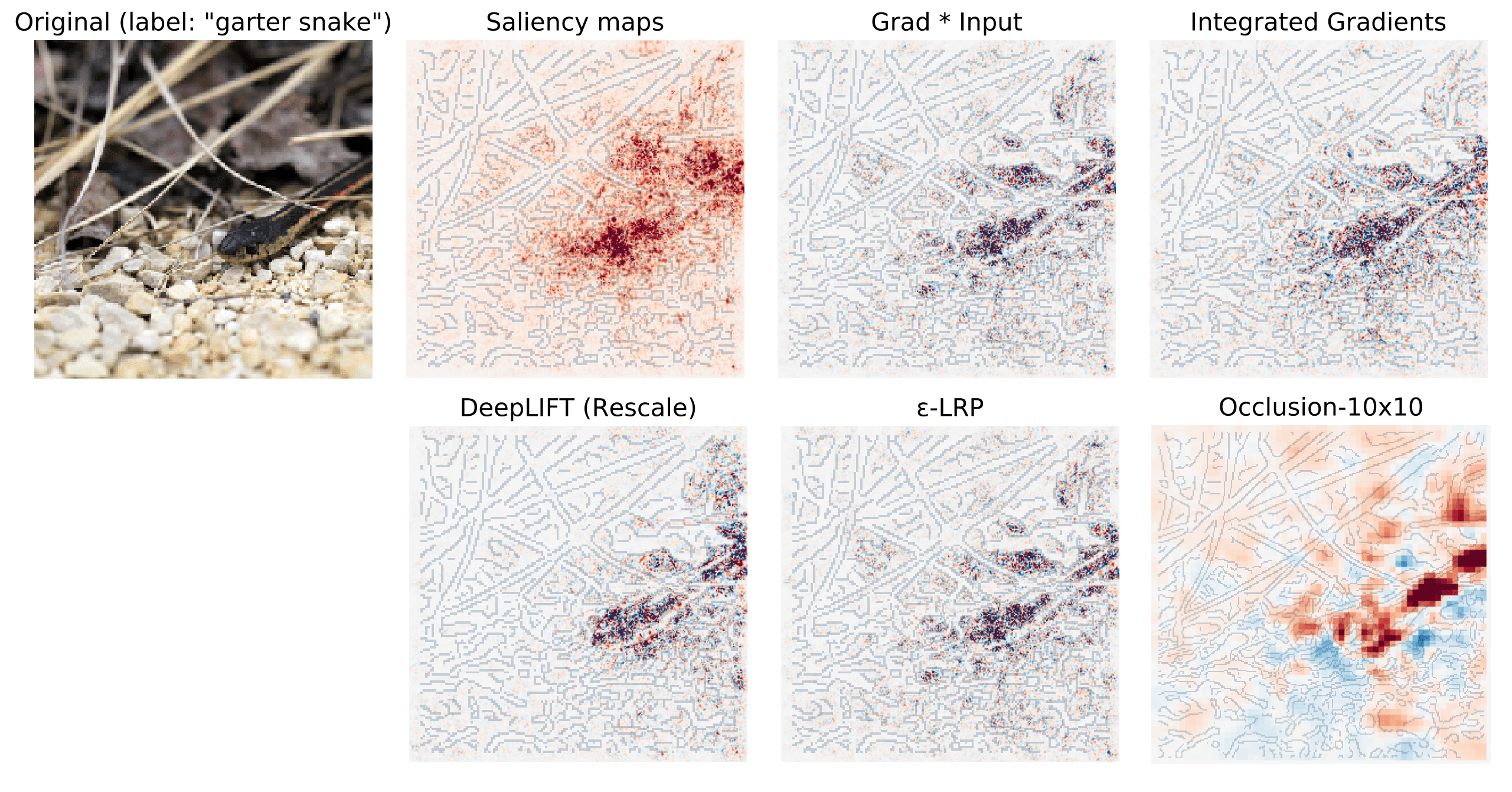}
    \caption{An overview of different explanation methods implemented in \cite{ancona2017towards, ancona2021towards}.
    The colours signify the areas of the image that each method considers as influential.
    Figure taken from \cite{ancona2021towards}.}
    \label{fig:DeepExplainComparison}
\end{figure}

Here, we show how attribution-based approaches can help to enable model transparency for ASR.
Focusing on the example of DeepSpeech, an end-to-end ASR model, we compare three different atribution methods:
Saliency Map \cite{simonyan2014deep}, Layer-wise Relevance Propagation (LRP) \cite{bach2015pixel}, and Shapley Additive Explanations (SHAP) \cite{castro2009polynomial,lundberg2017unified}.
The first two methods are based on the network's gradient towards the input, whereas the third is a game theory-based approach making use of so-called Shapley values \cite{shapley1997value}.
In the case of image classification, these methods consist of determining and visualising with a heat map regions of the input image that strongly influence a DNN's output (see Figure~\ref{fig:DeepExplainComparison}).
We show that an analogous visualization is possible for audio data and ASR.

The remainder of our paper is structured as follows.
In Section~\ref{sec:RelatedWorks}, we give a short general overview on explanation methods and on our chosen network, DeepSpeech.
We provide a more technical account of the three explanation techniques of interest and briefly outline our results in Section~\ref{sec:VisualizationTechniques}. 
In Section~~\ref{sec:Discussion}, we briefly introduce our computational experiments, followed by a discussion of the methods' advantages and shortcomings.
We conclude with an outlook in Section~\ref{sec:Conclusion}.

\section{Background on DeepExplain and DeepSpeech}\label{sec:RelatedWorks}

With neural networks becoming more and more complex, explanation or interpretation methods have become an important tool to gain insights on the networks' inner workings.
Despite possible shortcomings \cite{kindermans2019reliability, zhou2021feature}, these methods can provide valuable insights for a first understanding.
Further, explanation methods can also provide means to detect adversarial examples \cite{liu2020adversarial}.
There are different implementations for visualization techniques online, e.g., Captum \cite{kokhlikyan2020captum}.

For our implementations, we rely on DeepExplain\footnote{See ``https://github.com/marcoancona/DeepExplain''.}, a Tensorflow-based library that aims at creating a unified framework for several state-of-the-art explanation methods \cite{ancona2017towards}. 
Originally developed within the image domain, it has now been expanded to include techniques for NLP interpretation.
The methods implemented in DeepExplain are post-hoc attribution-based methods, that analyse the model after training (during inference) without modifying it. 
The methods include Saliency Map \cite{simonyan2014deep}, LRP \cite{bach2015pixel} and SHAP in a sampling-based approximation \cite{castro2009polynomial,lundberg2017unified,ancona2021towards}.
In our work, we have adapted these algorithms to work with DeepSpeech\footnote{See ``https://deepspeech.readthedocs.io/en'' and ``https://github.com/mozilla/DeepSpeech/''.}, an open-source end-to-end speech-to-text engine developed by Mozilla and based on Baidu's Deep Speech research paper \cite{hannun2014deep}.

In order to discuss the visualization, it is important to understand the rough architecture of DeepSpeech.
DeepSpeech's classification process can be divided into three stages: 
(i) a feature extraction stage that computes Mel-frequency cepstral coefficients (MFCCs); 
(ii) a long short-term memory (LSTM) neural network that takes MFCCs as input and outputs character probabilities; 
and (iii) a language model that turns the neural network's output into a properly formatted text. 

This paper focuses on the second step and visualize how the neural network gets to a frame's classification as a certain letter.
In the neural network, a pre-processing stage turns the input signal into an array of shape $(N, 26)$, where $N$ is the number of time frames (depending on input length) and $26$ is the number of considered Mel-frequency cepstral coefficients (MFCCs).
Each time frame is sequentially fed into the LSTM neural network along with the MFCCs of the nine previous and nine following time frames. 
The shape of the considered input for every step is thus $(19, 26)$.

\section{Visualization Techniques}\label{sec:VisualizationTechniques}
Among the wide range of interpretation methods developed for machine learning in the image domain, a prominent class is that of \emph{attribution methods}. 
An attribution quantifies the effect of each input feature, e.g., a pixel, on a neural network's output. 

A formal definition of attribution notion is provided in \cite{ancona2017towards}.
We adapt this for the case of ASR.
Given a NN that takes an input $x^i \in \mathbb{R}^{19 \times 26}$ (the currently evaluated audio slices) and produces an output $S(x^i) = [S_1(x^i),...,S_{28}(x^i)]$ (probability per letter), where $28$ is the total number of output neurons (representing the Latin alphabet including the space and hyphen symbols). 
The attribution for a specific target neuron $c$ is the matrix $A^i \in \mathbb{R}^{19 \times 26}$ of the contribution of each input feature $x^i_{j,k}$ to the output letter $S_c$. 
How the contribution $A^i_{j,k}$ is defined and computed depends on the chosen method. 
To allow the visualization of the decision process in DeepSpeech, we adapted two gradient-based methods from the image to the audio domain: Saliency Map \cite{simonyan2014deep} and LRP \cite{bach2015pixel}, and one game theory-based method: SHAP \cite{castro2009polynomial,lundberg2017unified}.
For each method, the attributions are defined as follows:
\paragraph*{Saliency Map}
    For a fixed time frame $i$, the contribution of each input feature $x^i_{j,k}$ is defined as the absolute value of the partial derivative of $S_c$ with respect to $x^i_{j,k}$, hence,
    \begin{align*} 
    A^i = \left|\frac{\partial S_c(x^i)}{\partial x^i}\right|.
    \end{align*}

\paragraph*{$\epsilon$ Layer-wise Relevance Propagation}
    The relevance is defined recursively by propagating backwards through the NN, from the output layer to the input layer.
    Specifically, let $R_u^{(l)}$ be the relevance of unit $u$ of layer $l$ with $l \in \left\{1, \, \ldots, \, L \right\}$ for a NN consisting of $L$ layers.
    Then, the relevance is first initialized in the last layer by:
    \begin{equation}
      R_u^{(L)} =
        \begin{cases}
          S_u(x^i), & \text{if $u$ is the target neuron,}\\
          0 & \text{else.}
        \end{cases} 
        \label{eq:lrp_eq1}
    \end{equation}

    The algorithm then iteratively redistributes the predictions from layer $l+1$ to layer $l$, until the first layer is reached.
    Let $z_{vu} = w^{(l+1, l)}_{vu} x_u^{(l)}$ be the weighted activation of a neuron $u$ from layer $l$ onto neuron $v$ in layer $l+1$, $b_v$ the additive bias of neuron $v$, and $\mathcal{N}_{l}$ the set of neurons in layer $l$, then we define the relevance as
    \begin{equation}
    R_u^{(l)} = \sum_{v \in \mathcal{N}_{l+1}} \frac{z_{vu}}{(1 + \epsilon) \sum_{u^\prime \in \mathcal{N}_{l}} (z_{vu^\prime} + b_v) } R_v^{(l+1)}.
    \label{eq:lrp_eq2}
    \end{equation}
    
    We set $\epsilon$ to $10^{-4}$, which is the default value used in \cite{ancona2021towards}.
    Note that the results may be highly sensitive to the chosen parameters, as $\epsilon$ absorbs relevance for weak or contradictory contributions to the activation of a neuron \cite{montavon2019layer}.

\paragraph*{Shapley values}
    In contrast to gradient-based methods, SHAP is a perturbation-based approach that estimates the attributions by performing $M$ random feature permutations. 
    More specifically, Shapley values \cite{shapley_values} are part of a concept from coalitional game theory which calculates a fairly distributed payout for each player of the game based on their contributions. 
    From a machine learning perspective, input features can be considered as ``players'', the model performing the prediction $f(\cdot)$ for a single input $x$ as a ``game'', and the difference between the actual prediction $f(x)$ and the average prediction for all instances $\mathop{\mathbb{E}(f(X))}$ as the ``gain''. 
    A coalition of multiple input features influence the final model output.
    Here, the Shapley values allow us to check which features have the most significant impact on the output by computing their marginal contributions. 
    For a single feature the Shapley value is the weighted sum of its marginal contributions. 
    The general definition of Shapley values for a given input feature $x^{i}_{j_1, k_1}$ at time step $i$ and input value $x^{i}$ is:

    \begin{align}
    \label{eq:shap_eq3}   
    \begin{split}
    \phi^{i}_{j_1, k_1}(x^{i}) = \sum_{S \subseteq \left\{ x^{i}_{j,k} \right\}_{j \neq j_1, k \neq k_1}} {\left[|S| \times \binom{F}{|S|}\right]}^{-1} \\ 
    \times \left[f_{\{ S \cup \{x^{i}_{j_1,k_1}\}\}}(x^{i}) - f_{\{S\}}(x^{i})\right],
    \end{split}
    \end{align}
where $|S|$ is the cardinality of the feature subset $S$ and $F$ is the total number of input features.
Hence in our experiments, $F = 19 \times 26$.
The marginal contribution of a feature $x^{i}_{j_1, k_1}$, calculated by $f_{\{ S \cup \{x^{i}_{j_1, k_1}\}\}}(x^{i}) - f_{\{S\}}(x^{i})$ is computed by subtracting the model predictions of two permutations constructed with and without $\{x^{i}_{j_1, k_1}\}$. 
Next, a weighting scheme described in \cite{lundberg2017unified} is applied to the marginal contributions. 
Finally, the weighted sum of marginal contributions is a Shapley value estimation $\phi^{i}_{j_1, k_1}(\cdot)$ of the input feature $x^{i}_{j_1, k_1}$.

As the total number of coalitions grows exponentially with the number of input features ($2^F$), the exact calculation of the Shapley values is computationally expensive.
Hence, to overcome this problem, we use Monte-Carlo sampling to estimate the Shapley values \cite{monte_carlo}. 
Additionally, to optimize the calculations we use background samples \cite{lundberg2017unified}, i.e., the median feature values of certain frames in the training data. 
Randomly chosen combinations of features from background samples are used to replace the features of the input sample which results into a random permutation. 
A feature-wise difference of two permutations is a marginal contribution of the feature $x^{i}_{j_1, k_1}$.
The sum of marginal contributions computed with the help of background samples is used in (\ref{eq:shap_eq3}) to estimate Shapley values for feature $x^{i}_{j_1, k_1}$.

\paragraph*{Displaying Audio Data}
In the original application of image classification, the attention values -- the values denoting the input's influence on the classification -- are plotted on top of the simplified image.
To transfer this visualization concept to the audio domain, we propose three different approaches.
Let $A$ denote the attribution matrix of size $(N, 19, 26)$, we consider:
\begin{itemize}
    \item \emph{Attribution per relative frame.} For a fixed relative frame $j \in \left\{1, \ldots, 19\right\}$, we can generate an attribution plot via $A^{i}_{j,k}$.
    When $j=9$, the influence of the current frame on its own classification is visualized.
    \item \emph{Attribution summed per frame.} For every frame, we can calculate its influence on the others' classification by summing diagonally (a zero is considered for out-of-range indexes): $\sum_{n=0}^{18} A^{i+n}_{j+n,k}$
    \item \emph{Attribution summed per window.} Centered at every frame $i$, we sum over all the attributions that the 19 frames have on $i^{\text{th}}$ frame's classification: $\sum_{j=1}^{19} A^i_{j,k}$.
\end{itemize}

In Figure \ref{fig:AttributionMethodsComparison}, we visualize the attributions using Saliency Map, LRP, and SHAP with the attribution summed per window.
It shows the contribution of each input feature to the predicted character indicated below the $x$ axis. 
Here, we consider the benign audio file for the recorded sentence \emph{``it has been mentioned but the article is not mine''}.
The gradient between two colors is used to express the impact of the attributions on the model prediction. 
Higher values, encoded in red, indicate a positive contribution of the features on the model's predictions. 
Features with negative impact on the decision process are instead encoded in blue.
We note that for an effective and informative plotting of the attributions, the color code needs to be chosen carefully, which we achieved by adapting the intensity scale for each method and choosing a fixed percentile value for all plots.

\section{Experiments and Discussion}\label{sec:Discussion}
In order to evaluate and compare the three visualization techniques, we have performed experiments covering 360 benign speech samples from the Mozilla Common Voice dataset\footnote{For more information, see ``https://commonvoice.mozilla.org/en/datasets''.}.
We have applied all three visualization techniques and the three different methods for displaying the data.
For the SHAP method, we have tried different background noises in order to compare the different attribution patterns.
Further, we have studied the behaviour on adversarial examples based on \cite{carlini2018audio} and \cite{qin2019imperceptible}, the latter adapted for DeepSpeech.

We aim to answer the following questions:
(1) Which attribution method seems the most suitable?
(2) What do we learn about the underlying ASR system?
We base our illustrations on one example.
According to our experiments, these findings generalize to the dataset considered.

To answer the first question, we start by comparing the two gradient-based methods, displayed in the two upper plots in Figure~\ref{fig:AttributionMethodsComparison} for the attributions summed per window and Figure~\ref{fig:SaliencyMaps_AttributionPerFrame} and Figure~\ref{fig:LRP_AttributionPerFrame} in the appendix for the attribution per frame.
For both, Saliency Map and LRP, the magnitude of the attribution is not evenly distributed amongst the frames.
We observe that frames nine to fourteen have a larger impact on the prediction than the other ones, especially the first eight frames. 
Hence, we report that later frames are more impactful for the final classification than earlier ones.
This may be due to the fact that DeepSpeech, being an LSTM model, already accounts for the past, so that the future provides more new information.
    
We further observe that the attribution is also unevenly distributed among the different MFCC bins. 
For LRP, the lowest coefficients are the most important ones, the vast majority of the attribution being shared between the first ten MFCCs. 
The result for the Saliency Map is quite different: 
the lowest and some of the highest MFCCs are the most relevant while coefficients five to twenty are less important.
We understand the different bins to follow from the fact that the LRP takes the input data into account by propagating the activation value of the neurons back through the network, whereas the Saliency Map method does not, as it instead simply computes the derivative of the NN's output. 


When comparing these results to the game theory inspired SHAP approach, we find very different results visualized in the bottom plot of Figure \ref{fig:AttributionMethodsComparison} and in Figure~\ref{fig:SHAP_AttributionPerFrame} in the appendix: 
(i) MFCC attributions obtained by applying the SHAP method have an equally spread magnitude response among all MFCC values and frames. 
(ii) We observe a clear difference in the response to frames containing letters compared to frames containing whitespace characters only. 

For observation (i), we see the following two reasons:
The background sample was chosen as the median of all frames with letters, i.e., input features were substituted with median values which affected the resulting MFCC attributions.
The large amount of input features affected the weighting coefficients and made them smaller which diminished the impact of each individual MFCC attribution.
The rationale behind observation (ii) might be that only frames corresponding to letters were used to construct the background sample. 
We randomly chose permutations of features in the input sample and substituted them by a random subset of frame features from the background sample.
For the background sample, we used the median values of the frames transcribed to letters.
According to our experiments, this allows a better Shapley estimation for actual letters. 

Overall, the permutation-based SHAP method provides an alternative look at the attributions of the audio samples by considering each frame and the according MFCC values as independent features.
In contrast, gradient-based methods additionally rely on the timestamp of the audio signal appearance. 
The SHAP method is model-agnostic, i.e., it applies to any method. 
Gradient-based methods (LRP and Saliency Map) can only be employed with speech recognition models trained using back propagation. 
The ability to incorporate the gradients for attribution computation renders these methods computationally much faster than perturbation-based methods, because the latter need a large number of permutations to achieve a good estimation of the attributions. 
Gradient and perturbation based methods can be used successfully to visualise the attributions of audio samples, answering different questions and providing different interpretations.

In summary, there is no visualization technique that is clearly the most recommendable. 
However, the LRP method seems preferable to us, as the method is fast to compute while providing very detailed information.
Additionally, since the different summation methods provide different insights on the frame's influence, they should all be considered.

To further shed light on the different visualization methods (second question) and to better understand DeepSpeech itself, we have further considered adversarial examples.
We assume that one can better detect adversarial examples (as a flaw of the NN) with a better understanding of the network.
Thus, we evaluated if our proposed visualization methods can be used to detect adversarial examples.
Visualizing attributions for adversarial examples computed using the method by \cite{carlini2018audio} and comparing them to benign audio samples showed clear differences. 
Generally speaking, adversarial examples tend to activate the NN to a greater extent, the magnitude of the attribution values is higher, especially for the highest MFCCs when we incorporate gradient-based visualisation methods. 
A similar effect was observed by applying the perturbation-based method but the difference between benign and adversarial samples was mainly captured among first $k$ frames. 
This phenomenon can be attributed to the fact that the used attack perturbs the whole audio sample uniformly and, hence, the largest difference can be measured at the very beginning of the audio sample where no signal is present in case of benign samples. 
With our first experiments at hand, we believe that these differences can be used to perform adversarial example detection making ASR systems more robust.

\section{Conclusion}\label{sec:Conclusion}
Visualization techniques in the audio domain are hard to evaluate: we do generally not know the ground truth \cite{zhang2019towards}, i.e., which features are relevant to the human and should be relevant for the classification system. 
Nevertheless, our proposed explanation methods offer practical insights, e.g., the computed attributions demonstrate which parts of the audio sample have the most significant impact on the speech model's prediction. 
Moreover, attribution methods help to get a deeper understanding of the behavior of ASR models. 
Finally, both, gradient and perturbation based methods, showed a great potential in resolving the problem of detecting audio adversarial samples. 
Future work may leverage our contribution to further explore this possibility in many domains, including psychology, security, and general machine learning:
How do humans process speech in comparison to the algorithm?
How do we need to adjust our ASR systems in order to be closer to the human hearing?
Can we use this to detect adversarial examples?

Answering these questions might help to find possible reasons for why neural networks can be fooled and thus allow developers to design strategies against adversarial attacks \cite{liu2020adversarial}.

\clearpage{}
\section{Acknowledgments}
This research was supported by the Bavarian Ministry of Economic Affairs, Regional Development and Energy.

The authors acknowledge N.~Schaaf for comments on the manuscript.

\bibliographystyle{IEEEtran}

\bibliography{sources.bib}

\section{Appendix}
\begin{figure*}[h!]
    \centering
    \includegraphics[width=0.9\textwidth]{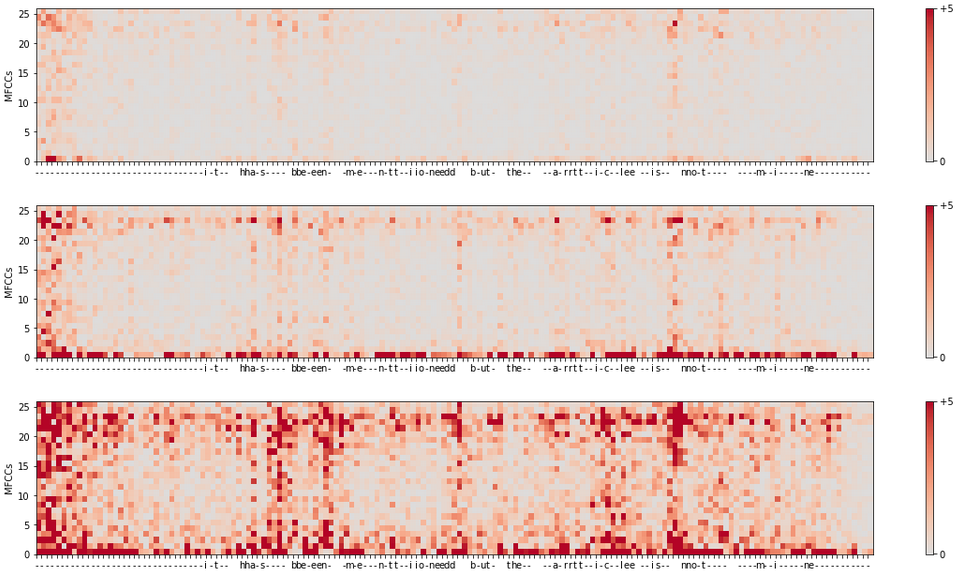}
    \caption{\textbf{Saliency Map} for the third, ninth and twelfth frame. 
    Later frames are assigned a higher attribution.
    Further, low and high MFFC values are assigned a higher attribution, if the frame itself is transcribed to a letter.
    For frames subscribed to a non-letter, the MFCCs' attribution is distributed more equally.}
    \label{fig:SaliencyMaps_AttributionPerFrame}
\end{figure*}

\begin{figure*}[h!]
    \centering
    \includegraphics[width=0.9\textwidth]{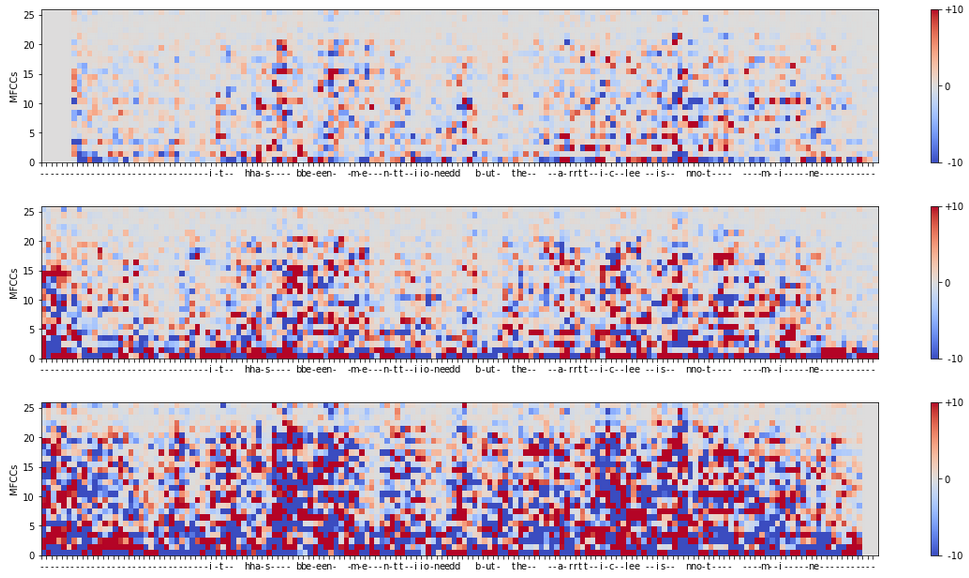}
    \caption{\textbf{LRP} for the third, ninth and twelfth frame.
    Later frames are assigned a higher attribution.
    Furher, low and mid-range MFCC values are assigned a higher attribution, if the frame itself is transcribed to a letter.
    For frames subscribed to a non-letter, the MFCCs' attribution is distrbuted more equally.}
    \label{fig:LRP_AttributionPerFrame}
\end{figure*}

\begin{figure*}[h!]
    \centering
    \includegraphics[width=0.9\textwidth]{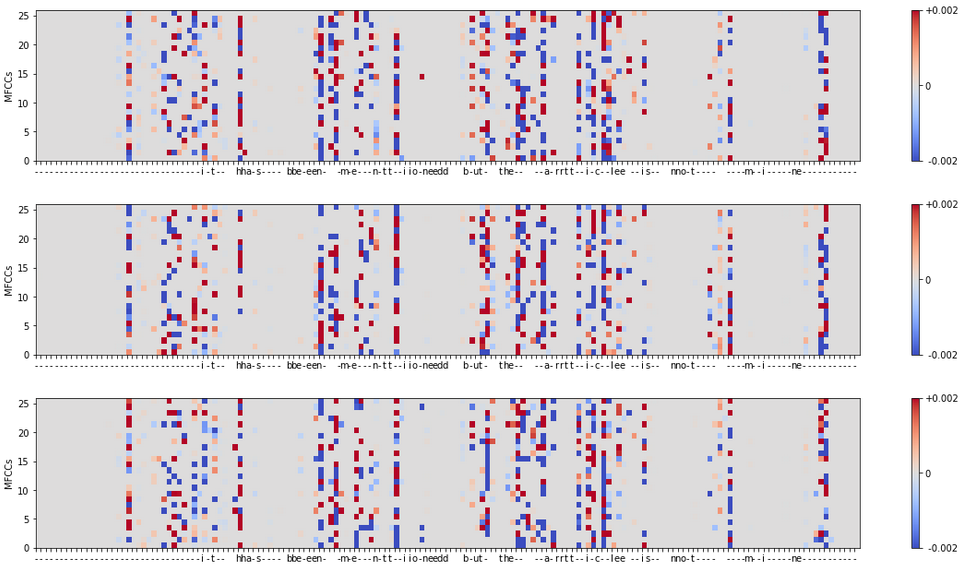}
    \caption{\textbf{SHAP} for the third, ninth and twelfth frame. All frames and MFCCs have similar attribution.}
    \label{fig:SHAP_AttributionPerFrame}
\end{figure*}

In this appendix, we provide additional figures for the same setting as in Figure~\ref{fig:AttributionMethodsComparison}. 
All plots show the third, the ninth, and the twelfth frame for the three different methods: Saliency Map (see Figure~\ref{fig:SaliencyMaps_AttributionPerFrame}), LRP (see Figure~\ref{fig:LRP_AttributionPerFrame}), and SHAP (see Figure~\ref{fig:SHAP_AttributionPerFrame}).

\end{document}